\documentclass[runningheads]{llncs}
\usepackage{graphicx}
\usepackage{comment}
\usepackage{amsmath,amssymb} % define this before the line numbering.
\usepackage{color}

\usepackage{fancyhdr}
\usepackage{multirow}
\usepackage{docmute}
\usepackage{chapterbib}

\newcommand{\tabincell}[2]{
\begin{tabular}{@{}#1@{}}#2\end{tabular}
}

\graphicspath{{figs/}}
\DeclareGraphicsExtensions{.pdf,.jpg,.jpeg,.png}

\begin{document}
% \linenumbers
\pagestyle{headings}
\mainmatter
\def\ECCVSubNumber{1369}  % Insert your submission number here

\title{Interpretable and Generalizable Person Re-Identification with Query-Adaptive Convolution and Temporal Lifting} % Replace with your title

\begin{comment}
\titlerunning{ECCV-20 submission ID \ECCVSubNumber}
\authorrunning{ECCV-20 submission ID \ECCVSubNumber}
\author{Anonymous ECCV submission}
\institute{Paper ID \ECCVSubNumber}
\end{comment}
%******************

%\begin{comment}
\titlerunning{ECCV 2020: Person Re-Identification with QAConv and TLift}
\author{Shengcai Liao$^1$\thanks{Corresponding Author.} and Ling Shao$^{1,2}$}
\authorrunning{Shengcai Liao and Ling Shao}
\institute{$^1$Inception Institute of Artificial Intelligence (IIAI), Abu Dhabi, UAE\\
$^2$Mohamed bin Zayed University of Artificial Intelligence, Abu Dhabi, UAE\\
\texttt{\email{\{scliao,ling.shao\}@ieee.org}}}
%\end{comment}
%******************
\maketitle
\thispagestyle{fancy}
\fancyhead{}
\chead{\small To appear in the European Conference on Computer Vision (ECCV), 2020}

\begin{figure}
\begin{minipage}{72mm}
\centering
\includegraphics[height=21mm]{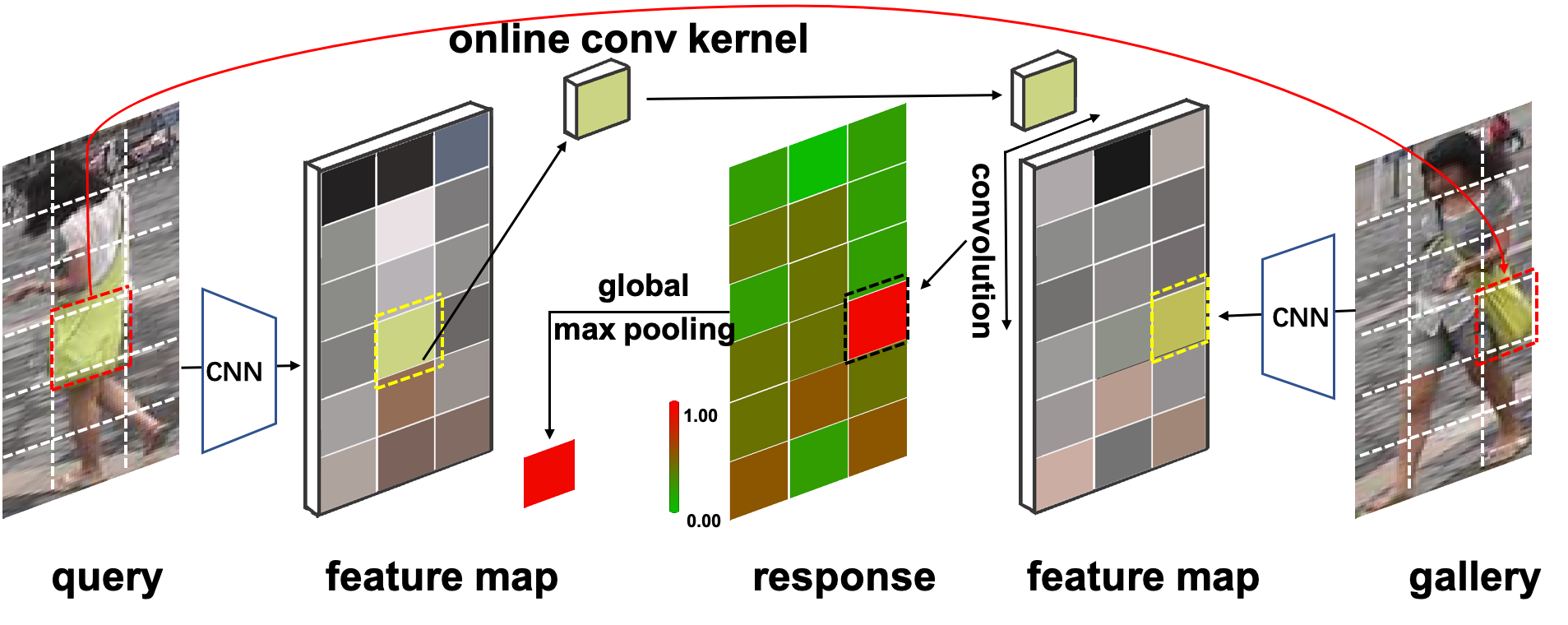}
\caption{QAConv constructs adaptive convolution kernels on the fly from query feature maps, and perform convolutions and max pooling on gallery feature maps to find the best local correspondences.}
\label{fig:qaconv}
\end{minipage}
\hspace{5mm}
\begin{minipage}{44mm}
\centering
\includegraphics[height=20mm]{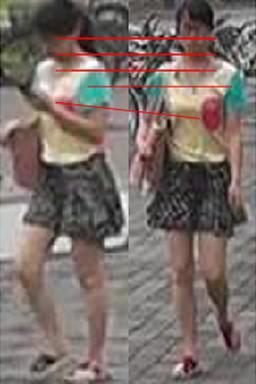}\hspace{1mm}
\includegraphics[height=20mm]{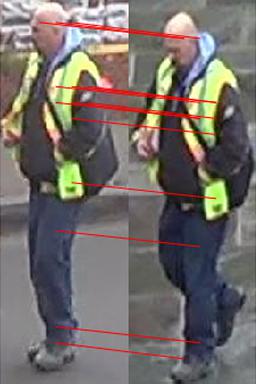}\hspace{1mm}
\includegraphics[height=20mm]{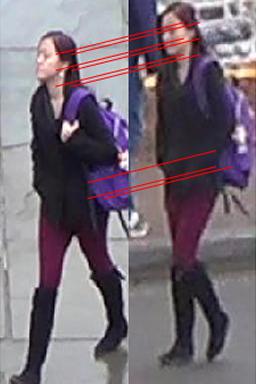}
\caption{Examples of the interpreted local correspondences from the outputs of the QAConv.}\label{fig:top-examples}
\end{minipage}
\end{figure}

\begin{abstract}
For person re-identification, existing deep networks often focus on representation learning. However, without transfer learning, the learned model is fixed as is, which is not adaptable for handling various unseen scenarios. In this paper, beyond representation learning, we consider how to formulate person image matching directly in deep feature maps. We treat image matching as finding local correspondences in feature maps, and construct query-adaptive convolution kernels on the fly to achieve local matching. In this way, the matching process and results are interpretable, and this explicit matching is more generalizable than representation features to unseen scenarios, such as unknown misalignments, pose or viewpoint changes. To facilitate end-to-end training of this architecture, we further build a class memory module to cache feature maps of the most recent samples of each class, so as to compute image matching losses for metric learning. Through direct cross-dataset evaluation, the proposed Query-Adaptive Convolution (QAConv) method gains large improvements over popular learning methods (about 10\%+ mAP), and achieves comparable results to many transfer learning methods. Besides, a model-free temporal cooccurrence based score weighting method called TLift is proposed, which improves the performance to a further extent, achieving state-of-the-art results in cross-dataset person re-identification. Code is available at \url{https://github.com/ShengcaiLiao/QAConv}.
%\keywords{We would like to encourage you to list your keywords within the abstract section}
\end{abstract}

%%%%%%%%% BODY TEXT
\section{Introduction}
Person re-identification is an active research topic in computer vision. It aims at finding the same person as the query image from a large volume of gallery images. With the progress in deep learning, person re-identification has been largely advanced in recent years. However, when generalization ability becomes an important concern, required by practical applications, existing methods usually lack satisfactory performance, evidenced by direct cross-dataset evaluation \cite{yi2014deep,Hu2014Cross}. To address this, many transfer learning, domain adaptation, and unsupervised learning methods, performed on the target domain, have been proposed. However, these methods require heavy computations in deployment, limiting their application in practical scenarios where the deployment machine may have limited resources to support deep learning and users may cannot wait for a time-consuming adaptation stage. Therefore, improving the baseline model’s generalization ability to support ready usage is still of urgent importance.

Most existing person re-identification methods compute a fixed representation vector, also known as a feature vector, for each image, and employ a typical distance or similarity metric (e.g. Euclidean distance or cosine similarity) for image matching. Without domain adaptation or transfer learning, the learned model is fixed as is, which is not adaptable for handling various unseen scenarios. Therefore, when generalization ability is a concern, it is expected to have an adaptive ability for the given model architecture.

In this paper, we focus on generalizable and ready-to-use person re-identification, through direct cross-dataset evaluation. Beyond representation learning, we consider how to formulate query-adaptive image matching directly in deep feature maps. %While how deep neural network works is still a mystery, we suppose that it does have strong representation capability in deep layers, though a black box, and we try to interpret its behaviour in the final image matching step.
Specifically, we treat image matching as finding local correspondences in feature maps, and construct query-adaptive convolution kernels on the fly to achieve local matching (see Fig. \ref{fig:qaconv}). In this way, the learned model benefits from adaptive convolution kernels in the final layer, specific to each image, and the matching process and result are interpretable (see Fig. \ref{fig:top-examples}), similar to traditional feature correspondence approaches \cite{SIFT,Bay-CVIU-08}. Probably because finding local correspondences through query-adaptive convolution is a common process among different domains, this explicit matching is more generalizable than representation features to unseen scenarios, such as unknown misalignments, pose or viewpoint changes. We call this Query-Adaptive Convolution QAConv. To facilitate end-to-end training of this architecture, we further build a class memory module to cache feature maps of the most recent samples of each class, so as to compute image matching losses for metric learning.

Through direct cross-dataset evaluation without further transfer learning, the proposed method achieves comparable results to many transfer learning methods for person re-identification. Besides, to explore the prior spatial-temporal structure of a camera network, a model-free temporal cooccurrence based score weighting method is proposed, named Temporal Lifting (TLift). %The basic idea is that nearby persons in one camera are potentially still nearby in another camera (see Fig. \ref{fig:method} (b)). Therefore, their corresponding matches in other cameras can serve as pivots to enhance the weight of their own nearby persons.
This is also computed on the fly for each query image, without statistical learning of a transition time model in advance. As a result, TLift improves person re-identification to a further extent, resulting in state-of-the-art results in cross-dataset evaluations.

To summarize, the novelty of this work include (i) a new deep image matching approach with query-adaptive convolutions, along with a class memory module for end-to-end training, and (ii) a model-free temporal cooccurrence based score weighting method. The advantages of this work are also two-fold. First, the proposed image matching method is interpretable, it is well-suited in handling misalignments, pose or viewpoint changes, and it also generalizes well in unseen domains. Second, both QAConv and TLift can be computed on the fly, and they are complementary to many other methods. For example, QAConv can serve as a better pre-trained model for transfer learning, and TLift can be readily applied by most person re-identification algorithms as a post-processing step.

\section{Related Works}
Deep learning approaches have largely advanced person re-identification in recent years~\cite{Ye2020Survey}. However, due to limited labeled data and a big diversity in real-world surveillance, %deep person re-identification
these methods usually have poor generalization ability in unseen scenarios. To address this, many unsupervised domain adaption (UDA) methods have been proposed~\cite{peng2016unsupervised,chang2019disjoint,wang2018transferable,li2018unsupervised,fan2018unsupervised,li2019unsupervised,Zhong-CVPR19-ECN,Yu-CVPR19-MAR,Yang-CVPR19-PAUL}, which show improved cross-dataset results than traditional methods, though requiring further training on the target domain. QAConv is orthogonal to transfer learning methods as it can provide a better baseline model for them (see Section \ref{subsec:sota} and Table \ref{tab:duke+market}).

There are many representation learning methods proposed to deal with viewpoint changes and misalignments in person re-identification, such as part-aligned feature representations~\cite{sun2017pcb,wang2018learning,suh2018part,zhao2017deeply}, pose-adapted feature representations~\cite{zhao2017spindle,saquib2018pose}, human parsing based representations~\cite{kalayeh2018human}, local neighborhood matching~\cite{ahmed2015improved,Li-CVPR-2014-DeepReID}, and attentional networks \cite{liu2017end,qian2017multi,liu2017hydraplus,xu2017jointly,si2018dual,li2018harmonious,xu2018attention}. While these methods present high accuracy when trained and tested on the same dataset, their generalization ability to other datasets is mostly unknown. Besides, beyond representation learning, QAConv focuses on image matching via local correspondences.

Generalizable person re-identification was first studied in our previous works \cite{yi2014deep,Hu2014Cross}, where direct cross-dataset evaluation was proposed. More recently, Song et al. \cite{song2019generalizable} proposed a domain-invariant mapping network by meta-learning, and Jia et al. \cite{jia2019frustratingly} applied the IBN-Net \cite{pan2018two} to improve generalizability, while QAConv is preliminarily reported in \cite{Liao-arXiv2019-QAConv}. QAConv is orthogonal to methods of network design, for example, it can also be applied on the IBN-Net for improvements.

For deep feature matching, Kronecker-Product Matching (KPM) \cite{shen2018end} computes a cosine similarity map by outer product for softly aligned element-wise subtraction. Besides, Bilinear Pooling \cite{Lin2015Bilinear,Ustinova2017bilinear,suh2018part-aligned} and Non-local Neural Networks \cite{wang2018non} also apply the outer product for part-aligned or self-attended representation learning. Different to the above methods, QAConv is a convolutional matching method but not simply outer product especially when its kernel size $s>1$. It is explicitly designed for local correspondence matching, interpretation, and generalization, in a straightforward way without other branches. %We explain how and why it works, and especially focus on generalizability.

For post-processing, re-ranking is a technique of refining matching scores, which further improves person re-identification~\cite{liu2013pop,yu2017divide,zhong2017re,saquib2018pose}. Besides, temporal information is also a useful cue to facilitate cross-camera person re-identification~\cite{lv2018unsupervised,wang2019spatial-temporal}. While existing methods model transition times across different cameras but encounter difficulties in complex transition time distributions, the proposed TLift method applies cooccurrence constraint within each camera to avoid estimating transition times, and it is model-free and can be computed on the fly.

For memory based loss, ECN \cite{Zhong-CVPR19-ECN} proposed an exemplar memory which caches feature vectors of every instance for UDA. This makes the instance-level label inference convenient but limits its scalability. In contrast, class memory is independently designed \cite{Liao-arXiv2019-QAConv}, which is more efficient working in class level.

\section{Query-adaptive Convolution}
\subsection{Query-adaptive Convolutional Matching}
For face recognition and person re-identification, most existing methods do not explicitly consider the relationship between two input images under matching, but instead, like classification, they treat each image independently and apply the learned model to extract a fixed feature representation. Then, image matching is simply a distance measure between two representation vectors, regardless of the direct relationship between the actual contents of the two images.

In this paper, we consider the relationship between two images, and try to formulate adaptive image matching directly in deep feature maps. Specifically, we treat image matching as finding local correspondences in feature maps, and construct query-adaptive convolution kernels on the fly to achieve local matching. As shown in Fig. \ref{fig:qaconv} and Fig. \ref{fig:arch}, to match two images, each image is firstly fed forward into a backbone CNN, resulting in a final feature map of size $[1, d, h, w]$, where $d$ is the number of output channels, and $h$ and $w$ are the height and width of the feature map, respectively. Then, the channel dimension of both feature maps is normalized by the $\ell2$-norm. After that, local patches of size $[s, s]$  at every location of the query feature map are extracted, and then reorganized into $[hw, d, s, s]$ as a convolution kernel, with input channels $d$, output channels $hw$, and kernel size $[s, s]$. This acts as a query-adaptive convolution kernel, with parameters constructed on the fly from the input, in contrast to fixed convolution kernels in the learned model. Upon this, the adaptive kernel can be used to perform a convolution on another feature map, resulting in $[1, hw, h, w]$ similarities.

Since feature channels are $\ell2$-normalized, when $s=1$, the convolution in fact measures the cosine similarity at every location of the two feature maps. %This special case is similar to one layer of KPM \cite{shen2018end}.
Besides, since the convolution kernel is adaptively constructed from the image content, these similarity values exactly reflect the local matching results between the two input images. Therefore, an additional global max pooling (GMP) operation will output the best local matches, and the maximum indices found by GMP indicate the best locations of local correspondences, which can be further used to interpret the matching result, as shown in Fig. \ref{fig:top-examples}. %Note that, by matching only two images, the above process can also be done by matrix multiplication. However, when a batch of images is considered, a convolution is more suited and more efficient. Also note
Note that GMP can also be done along the $hw$ axis of the $[1, hw, h, w]$ similarity map. That is, seeking the best matches can be carried out from both sides of the images. Concatenating the output will result in a similarity vector of size $2hw$ for each pair of images.

\subsection{Network Architecture}
\begin{figure}
\begin{minipage}{58mm}
\centering
\includegraphics[width=58mm]{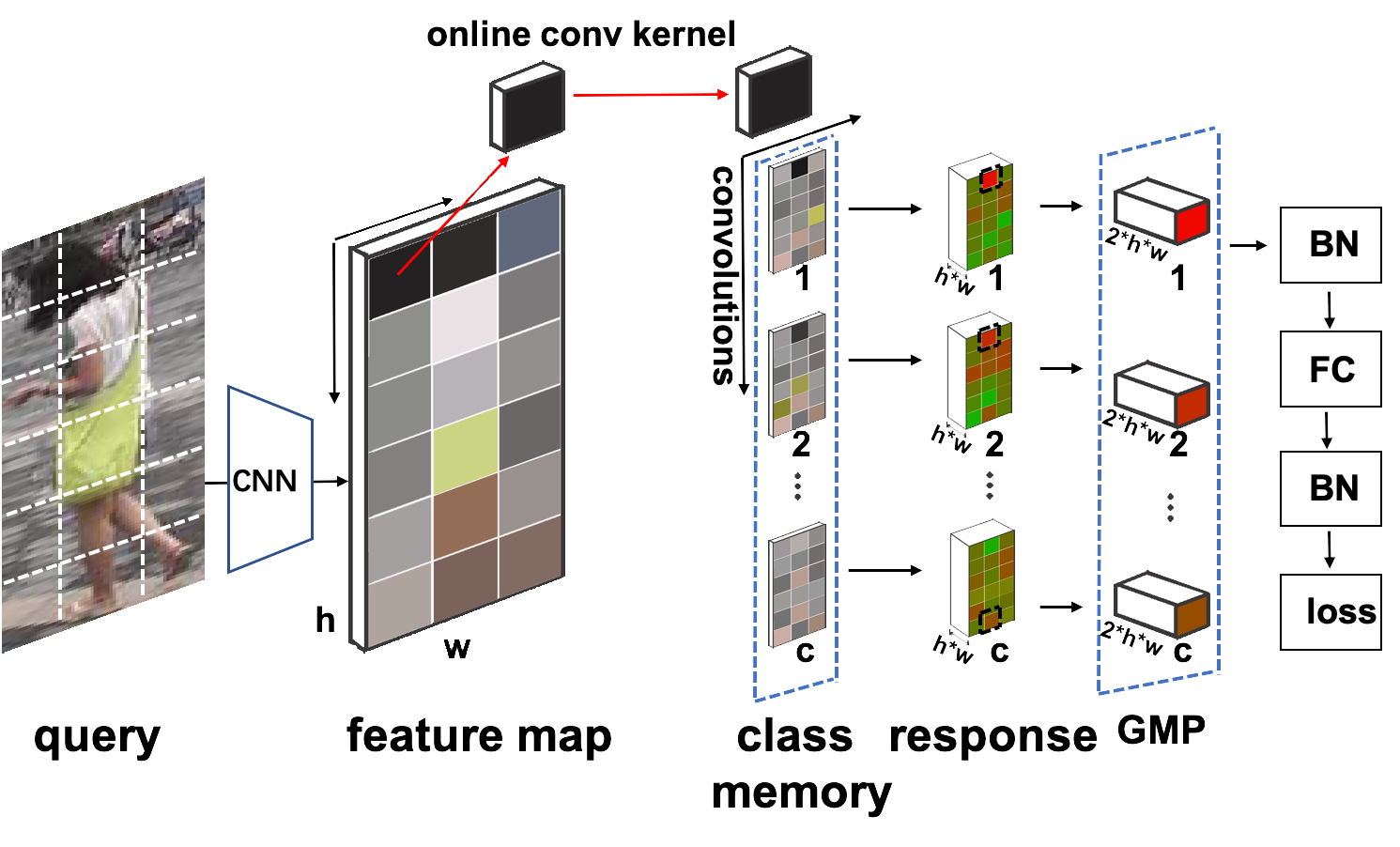}
\caption{Architecture of the QAConv. GMP: global max pooling. BN: batch normalization. FC: fully connection.}
\label{fig:arch}
\end{minipage}
\hspace{4mm}
\begin{minipage}{58mm}
\centering
\includegraphics[width=58mm]{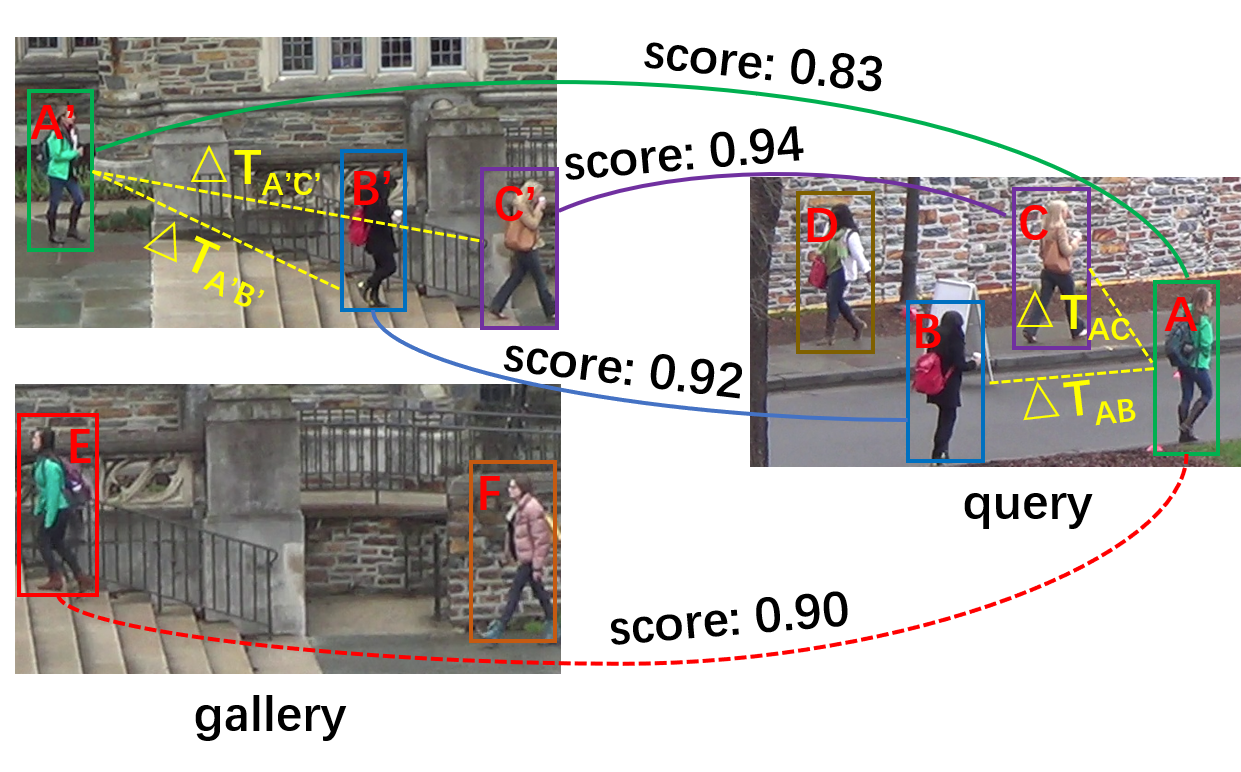}
\caption{Illustration of the proposed TLift approach.}% $A$ is the query person. $E$ is more similar than $A'$ to $A$ in another camera. With nearby persons $B$ and $C$, and their cross-camera top retrievals $B'$ and $C'$ acting as pivots, the score of $A'$ near $B'$ and $C'$ can be temporally lifted, while the score of $E$ will be reduced due to no such pivot.}
\label{fig:TLift}
\end{minipage}
\end{figure}
The architecture of the proposed query-adaptive convolution method is shown in Fig. \ref{fig:arch}, which consists of a backbone CNN, the QAConv layer for local matching, a class memory layer for training, a global max pooling layer, a BN-FC-BN block, and, finally, a similarity output by a sigmoid function for evaluation in the test phase or loss computation in the training phase. %The input sizes of the BN-FC-BN block are all $2\times h\times w$.
The output size of the FC layer is $1$, which acts as a binary classifier or a similarity metric, indicating whether or not one pair of images belongs to the same class. The two BN (batch normalization \cite{Ioffe-BatchNorm-ICML15}) layers are all one-dimensional. They are used to normalize the similarity output and stabilize the gradient during training.

\subsection{Class Memory and Update}
%
%To train the QAConv image matching architecture, we need to form sufficient training image pairs. A natural way to do this is to use mini batches for training, and form image pairs within each mini batch. However, this is not efficient for sampling the whole training set and the convergence is slow, since there are $N^2$ possible combinations of all sample pairs, where $N$ is the number of training images. Therefore, we
We propose a class memory module to facilitate the end-to-end training of  the QAConv network. Specifically, a $[c, d, h, w]$ tensor buffer is registered, where $c$ is the number of classes. For each mini batch of size $b$, %after the loss computation (introduced below),
the $[b, d, h, w]$ feature map tensor of the mini batch will be updated into the memory buffer. We use a direct assignment update strategy, that is, each $[1, d, h, w]$ sample of class $i$ from the mini batch will be assigned into location $i$ of the $[c, d, h, w]$ memory buffer.

An exponential moving average update can also be used here. However, in our experience this is inferior to the direct replacement update. There might be two reasons for this. First, the replacement update caches feature maps of the most recent samples of each class, so as to reflect the most up-to-date state of the current model for loss computation. Second, since our task is to carry out image matching with local details in feature maps for correspondences, exponential moving average may smooth the local details of samples from the same class.% Note that recently an exemplar memory module is also proposed in ECN \cite{Zhong-CVPR19-ECN}. In contrast to class memory, exemplar memory caches feature vectors of every instance for unsupervised transfer learning, where KNN is applied for pseudo label assignment, and exemplar memory makes the instance-level KNN convenient.

\subsection{Loss Function}
With a mini batch of size $[b, d, h, w]$ and class memory of size $[c, d, h, w]$, $b\times c$ pairs of similarity values will be computed by QAConv after the BN-FC-BN block. We use a sigmoid function to map the similarity values into $[0,1]$, and compute the binary cross entropy loss. Since there are far more negative than positive pairs, to balance them and enable online hard example mining, we apply the focal loss \cite{lin2017focal} to weight the binary cross entropy. That is,
\begin{equation}\label{eq:loss}
\ell(\theta) = -\frac{1}{b}\sum_{i=1}^{b}\sum_{j=1}^{c}(1-\hat{p}_{ij}(\theta))^{\gamma}log(\hat{p}_{ij}(\theta)),\\
\end{equation}
where $\theta$ is the network parameter, %$\gamma$ is the focusing parameter (by default $\gamma=2$ as in \cite{lin2017focal}), and
$\gamma=2$ is the focusing parameter \cite{lin2017focal}, and
\begin{equation}
\begin{aligned}
&\hat{p}_{ij}=
\begin{cases}
p_{ij}  & \mbox{if $y_{ij}=1$,}\\
1-p_{ij}& \mbox{otherwise,}
\end{cases}
\end{aligned}
\end{equation}
where $y_{ij}=1$ indicates a positive pair, while a negative pair otherwise, and $p_{ij}\in[0,1]$ is the sigmoid probability.

\section{Temporal Lifting}
For person re-identification, to explore the prior spatial-temporal structure of a camera network, usually a transition time model is learned to measure the transition probability. However, for a complex camera network and various person transition patterns, it is not easy to learn a robust transition time distribution. In contrast, in this paper a model-free temporal cooccurrence based score weighting method is proposed, which is called Temporal Lifting (TLift). TLift does not model cross-camera transition times which could be variable and complex. Instead, TLift makes use of a group of nearby persons in each single camera, and find similarities between them.

Fig. \ref{fig:TLift} illustrates the idea. A basic assumption is that people nearby in one camera are likely still nearby in another camera. Therefore, their corresponding matches in other cameras can serve as pivots to enhance the weights of other nearby persons. In Fig. \ref{fig:TLift}, $A$ is the query person. $E$ is more similar than $A'$ to $A$ in another camera. With nearby persons $B$ and $C$, and their top retrievals $B'$ and $C'$ acting as pivots, the matching score of $A'$ can be temporally lifted since it is a nearby person of $B'$ and $C'$, while the matching score of $E$ will be reduced since there is no such pivot.

Formally, suppose $A$ is the query person in camera $Q$, then, the set of nearby persons to $A$ in camera $Q$ is defined as $R = \{B | \Delta T_{AB} < \tau, \forall B \in Q\}$,
%
%\begin{equation}
%R = \{B | \Delta T_{AB} < \tau, \forall B \in Q\},
%\end{equation}
%
where $\Delta T_{AB}$ is the within-camera time difference between persons $A$ and $B$, and $\tau$ is a threshold on $\Delta T$ to define nearby persons. Then, for each person in $R$, cross-camera person retrieval will be performed on a gallery camera $G$ by QAConv or other methods, and the overall top K retrievals for $R$ are defined as the pivot set $P$. Next, each person in $P$ acts as an ensemble point for 1D kernel density estimation on within-camera time differences in $G$, and the temporal matching probability between $A$ and any person $X$ in camera $G$ will be computed as
\begin{equation}
p_t(A,X) = \frac{1}{|P|}\sum_{B\in P}e^{-\frac{\Delta T_{BX}^2}{\sigma^2}},
\end{equation}
where $\sigma$ is the sensitivity parameter of the time difference. Then, this temporal probability is used to weight the similarity score of appearance models using a multiplication fusion as $p(A,X) = (p_t(A,X) + \alpha)p_a(A,X)$,
%
%\begin{equation}
%p(A,X) = (p_t(A,X) + \alpha)p_a(A,X),
%\end{equation}
%
where $p_a(A,X)$ is the appearance based matching probability (e.g. by QAConv), and $\alpha$ is a regularizer.

This way, true positives near pivots will be lifted, while hard negatives far from pivots will be suppressed. Note that this is also computed on the fly for each query image, without learning a transition time model in advance. Therefore, it does not require training data, and can be readily applied by many other person re-identification methods.

\section{Experiments}
\subsection{Implementation Details}
The proposed method is implemented in PyTorch, based upon an adapted version \cite{zhong2018camstyle} of the open source person re-identification library (open-reid)\footnote{https://cysu.github.io/open-reid/}. Person images are resized to $384\times128$. The backbone network is the ResNet-152 \cite{he2016resnet} pre-trained on ImageNet, unless otherwise stated. The layer3 feature map of the backbone network is used, since the size of the layer4 feature map is too small. A $1\times1$ convolution with 128 channels is further appended to reduce the final feature map size. The batch size of samples for training is 32. The SGD optimizer is applied, with a learning rate of 0.001 for the backbone network, and 0.01 for newly added layers. They are decayed by 0.1 after 40 epochs, and the training stops at 60 epochs. The whole QAConv is end-to-end jointly trained, while class memory is updated only after the loss computation. Considering the memory consumption and the efficiency, the kernel size of QAConv is set to $s=1$. Parameters for TLift are $\tau=100$, $\sigma=200$, $K=10$, and $\alpha=0.2$. They are not sensitive in a broad range, as analyzed in the Appendix.

% A Random Occlusion (RO) module is implemented for data augmentation of the QAConv training to simulate person occlusions, which is very similar to the cutout \cite{devries2017improved} and Random Erasing (RE) \cite{zhong2020random} methods. In the RE implementation, the target erasing area is sampled from a combination of random area and aspect ratio, which could exceed the original image height or width. Therefore, it needs to try multiple times (100 by default) to generate a reasonable region for erasing. In contrast, in our implementation of the RB module, a square block is used, with the size randomly sampled at most $0.8\times width$ of the image. Then the square block is filled with white pixels. Note that with a simple square block, there is no need to sample multiple times of areas and aspect ratios and check the validity, and hence the generation process is more efficient. Beyond this, only a random horizontal flipping is used for data augmentation.

A random occlusion module is implemented for data augmentation, which is similar to the random erasing \cite{zhong2020random} and cutout \cite{devries2017improved} methods (see Appendix for comparisons). Specifically, a square area is generated with the size randomly sampled at most $0.8\times width$ of the image. Then this square area is filled with white pixels. It is useful for QAConv because random occlusion forces QAConv to learn various local correspondences, instead of only saliency but easy ones. Beyond this, only a random horizontal flipping is used for data augmentation.

\subsection{Datasets}
Experiments were conducted on four large person re-identification datasets, Market-1501 \cite{zheng2015smarket}, DukeMTMC-reID \cite{ristani2016duke,zheng2017unlabeled}, CUHK03 \cite{Li-CVPR-2014-DeepReID}, and MSMT17~\cite{Wei-CVPR18-PTGAN}. The Market-1501 dataset contains 32,668 images of 1501 identities captured from 6 cameras. There are 12,936 images from 751 identities for training, and 19,732 images from 750 identities for testing%, which is further divided into 3,368 query images and 15,913 gallery images
. The DukeMTMC-reID is a subset of the multi-target and multi-camera pedestrian tracking dataset DukeMTMC~\cite{ristani2016duke}. It includes 1,812 identities and 36,411 images, where 16,522 images of 702 identities are used for training, and the remainings for test. The CUHK03 dataset includes 13,164 images of 1,360 pedestrians. We adopted the CUHK03-NP protocol provided in \cite{zhong2017re}, where images of 767 identities were used for training, and other images of 700 identities were used for test. Besides, we used the detected subset for evaluation, which is more challenging. The MSMT17 dataset is the largest person re-identification dataset to date, which contains 4,101 identities and 126,441 images captured from 15 cameras. It is divided into a training set of 32,621 images from 1,041 identities, and a test set with the remaining images from 3,010 identities. %The test set further contains 11,659 query images and 82,161 gallery images.

Cross-dataset evaluation was performed in these datasets, by training on the training subset of one dataset (except that in MSMT17 we used all images for training following \cite{Yu-CVPR19-MAR,Yang-CVPR19-PAUL}), and evaluating on the test subset of another dataset. The cumulative matching characteristic (CMC) and mean Average Precision (mAP) were used as the performance evaluation metrics. All evaluations followed the single-query evaluation protocol.

The Market-1501 and DukeMTMC-reID datasets are with frame numbers available, so that it is able to evaluate the proposed TLift method. The DukeMTMC-reID dataset has a good global and continuous record of frame numbers, and it is synchronized by providing offset times. In contrast, the Market-1501 dataset has only independent frame numbers for each session of videos from each camera. %For several sessions from each camera, we roughly calculated the overall time frames of each session as offset, and made a cumulative record by assuming the video sessions were continuously recorded.
Accordingly we simply made a cumulative frame record by assuming continuous video sessions. After that, frame numbers were converted to seconds in time by dividing the Frames Per Second (FPS) in video records, where FPS=59.94 for the DukeMTMC-reID dataset and FPS=25 for the Market-1501 dataset.

\subsection{Ablation Study}
Some ablation studies have been conducted to understand the proposed method, in the context of direct cross-dataset evaluation between the Market-1501 and DukeMTMC-reID datasets.
%First, the role of the random block (RB) implementation as data augmentation is evaluated, compared to the random erasing (RE) method of \cite{zhong2017random}. From the results shown in Table \ref{tab:block}, it can be observed that the new implementation of the RB module performs better than the original RE implementation, as well as a baseline without RE or RB. This may be because the restricted square shape occludes person bodies within a local area. Besides, intuitively, RE/RB is useful for QAConv because random occlusion forces QAConv to learn various local correspondences, instead of only saliency but easy ones. Therefore, considering also the efficiency of the RB implementation, the new implementation is preferred in the training of the proposed QAConv algorithm.
%
%\begin{table}
%    \centering
%    \caption{Role of random block.}\label{tab:block}
%    \begin{tabular}{|l|c|c|c|c|}
%      \hline
%      \multirow{2}{*}{\tabincell{c}{Method}}  & \multicolumn{2}{|c|}{Market$\rightarrow$Duke} &  \multicolumn{2}{|c|}{Duke$\rightarrow$Market} \\
%      \cline{2-5}
%        & Rank-1 & mAP  & Rank-1 & mAP \\
%      \hline
%      QAConv without RE/RB & 50.5 & 29.5 & 61.6 & 28.4\\
%      QAConv with RE~\cite{zhong2017random} & 51.6 & 30.6 & 62.0 & 29.8\\
%      QAConv with RB & \textbf{54.4} & \textbf{33.6} &\textbf{62.8} &\textbf{31.6}\\
%      \hline
%    \end{tabular}
%\end{table}
%
%
First, to understand the QAConv loss, several other loss functions, including the classical softmax based cross entropy loss, the center loss~\cite{Wen-ECCV16-CenterLoss,Jin-IJCB17-CenterLossReID}, the Arc loss (derived from the ArcFace method~\cite{Deng-CVPR19-ArcFace} which is effective for face recognition), and the proposed class memory based loss, are evaluated for comparison. For these compared loss functions, the global average pooling of layer4 (better than layer3) of the ResNet-152 is used for feature representation, and the cosine similarity measure is adopted instead of the QAConv similarity. For the class memory loss, feature vectors are cached in memory instead of learnable parameters, and the same BN layer and Eq. (\ref{eq:loss}) are applied after calculating the cosine similarity values between mini-batch features and memory features.

From results shown in Table \ref{tab:loss}, it is obvious that QAConv improves existing loss functions by a large margin, with 13.7\%-19.5\% improvements in Rank-1, and 9.6\%-11.1\% in mAP. Interestingly, large margin classifiers improves the softmax cross-entropy baseline when trained on the Market-1501 dataset, but do not have such improvements when trained on DukeMTMC-reID. This is probably due to many ambiguously labeled or closely walking persons in DukeMTMC-reID (see Section \ref{sec:discuss}), which may confuse the strict large margin training. Note that the class memory based loss only performs comparable to other existing losses, indicating that the large improvement of QAConv is mainly due to the new matching mechanism, rather than the class memory based loss function. Besides, the Arc loss published recently is one of the best face recognition method, but it does not seem to be powerful when applied in person re-identification\footnote{We have tried different hyper parameters and reported the best results. The best margin values were found to be 0.5 on Market-1501 and 0.2 on DukeMTMC-reID.}. In our experience, the choice of loss functions does not largely influence person re-identification performance. Similar as in face recognition, existing studies \cite{Wen-ECCV16-CenterLoss,Deng-CVPR19-ArcFace} show that new loss functions do have improvements, but cannot be regarded as significant ones over the softmax cross entropy baseline. Therefore, we may conclude that the large improvement observed here is due to the new matching scheme, instead of different loss configurations (see Appendix for more analyses).
\begin{table}
    \centering
    \caption{Role of loss functions (\%).}\label{tab:loss}
    \begin{tabular}{|c|c|c|c|c|}
      \hline
      \multirow{2}{*}{\tabincell{c}{Method}}  & \multicolumn{2}{|c|}{Market$\rightarrow$Duke} &  \multicolumn{2}{|c|}{Duke$\rightarrow$Market} \\
      \cline{2-5}
        & Rank-1 & mAP  & Rank-1 & mAP \\
      \hline
      Softmax cross-entropy & 34.9 & 18.4 & 48.5 & 21.4\\
      Arc loss~\cite{Deng-CVPR19-ArcFace} & 35.3 & 17.1 & 48.9 & 21.4\\
      Center loss~\cite{Wen-ECCV16-CenterLoss,Jin-IJCB17-CenterLossReID} & 38.9 & 22.1 & 48.8 & 22.0\\
      Class memory loss & 40.7 & 21.8 & 47.8 & 20.5\\
      QAConv & \textbf{54.4} & \textbf{33.6} & \textbf{62.8} & \textbf{31.6}\\
      \hline
    \end{tabular}
\end{table}

Next, to understand the role of re-ranking (RR), the k-reciprocal encoding method \cite{zhong2017re} is applied upon QAConv. From results shown in Table \ref{tab:tlift}, it can be seen that enabling re-ranking do improve the performance a lot, especially with mAP, which is increased by 18.8\% under Market$\rightarrow$Duke, and 19.6\% under Duke$\rightarrow$Market. This improvement is much more significant based on QAConv than that based on other methods as reported in \cite{zhong2017re}. This is probably because the new QAConv matching scheme better measures the similarity between images, which benefits the reverse neighbor based re-ranking method.% Note that the large improvement in mAP is important for the subsequent TLift, because it is relied on cross-camera top-k retrievals.

Furthermore, based on QAConv and re-ranking, the contribution of TLift is evaluated, compared to a recent method called TFusion (TF) \cite{Lv18-TFusion}, which is originally designed to iteratively improve transfer learning. From results shown in Table \ref{tab:tlift}, it can be observed that employing TLift to explore temporal information further improves the results, with Rank-1 improved by 8.2\%-10.2\%, and mAP by 7.0\%-8.8\%. This improvement is complementary to re-ranking, so they can be combined. As for the existing method TFusion, it appears to be not stable, as a large improvement can be observed under Market$\rightarrow$Duke, but little improvement can be obtained under Duke$\rightarrow$Market, or even the mAP is clearly decreased\footnote{Note that TFusion parameters were optimized on each dataset to get the best results, but for TLift we used fixed parameters for all datasets (see Appendix for analysis).}. This may be because TFusion is based on learning transition time distributions across cameras, which is not easy to deal with complex camera networks and person transitions as in the Market-1501 (various repeated presences per person in one camera). In contrast, the TLift method only depends on single-camera temporal information which is relatively more easy to handle. Note that TLift can also be generally applied to other methods for improvements, as shown in the Appendix. Besides, as shown in Table \ref{tab:tlift}, directly applying TLift to QAConv without re-ranking also improves the performance a lot.
\begin{table}
    \centering
    \caption{Performance (\%) of different post-processing methods.}\label{tab:tlift}
    \begin{tabular}{|l|c|c|c|c|}
      \hline
      \multirow{2}{*}{\tabincell{c}{Method}}  & \multicolumn{2}{|c|}{Market$\rightarrow$Duke} &  \multicolumn{2}{|c|}{Duke$\rightarrow$Market} \\
      \cline{2-5}
        & Rank-1 & mAP  & Rank-1 & mAP \\
      \hline
      QAConv & 54.4 & 33.6 & 62.8 & 31.6 \\
      QAConv + TLift & 62.7 & 45.3 & 61.5 & 40.6\\
      QAConv + RR~\cite{zhong2017re} & 61.8 & 52.4 & 68.5 & 51.2 \\
      QAConv + RR + TF~\cite{Lv18-TFusion} & \textbf{70.7} & \textbf{61.9} & 68.6 & 47.2\\
      QAConv + RR + TLift & 70.0 & 61.2 & \textbf{78.7} & \textbf{58.2} \\
      \hline
    \end{tabular}
\end{table}

Finally, to understand the effect of the backbone network, the QAConv results with the ResNet-50 as backbone are also reported in Tables \ref{tab:duke+market} and \ref{tab:cuhk+msmt}, compared to the default ResNet-152 (denoted as QAConv$_{50}$ and QAConv$_{152}$, respectively). As can be observed, a larger network ResNet-152 does have a better performance due to its larger learning capability. It can improve the Rank-1 accuracy over the QAConv$_{50}$ by 1.3\%-7.3\%, and the mAP by 0.8\%-5.5\%. Besides, there are also consistent improvements in case of combining re-ranking and TLift. Hence, it seems that this larger network, which contains more learnable parameters, does not have the overfitting problem when equipped with QAConv. Note that, though ResNet-152 is a very large network requiring heavy computation, in practice, it can be efficiently reduced by knowledge distillation \cite{hinton2015distilling}.

\subsection{Comparison to the State of the Arts}\label{subsec:sota}
There are a great number of person re-identification methods since this is a very active research area. Here we only list recent results for comparison due to limited space. The cross-dataset evaluation results on the four datasets are listed in Tables \ref{tab:duke+market} and \ref{tab:cuhk+msmt}. Considering that many person re-identification methods employ the ResNet-50 network, for a fair comparison, the following analysis is based on the QAConv$_{50}$ results. Note that this paper mainly focuses on cross-dataset evaluation. Therefore, some recent methods performing unsupervised learning on the target dataset are not compared here, such as the TAUDL~\cite{li2018unsupervised}, UTAL~\cite{li2019unsupervised}, and UGA~\cite{Wu-CVPR19-UGA}, and also partially due to the fact that they use single-camera target identity labels for training. There are mainly two groups of methods listed in Table \ref{tab:duke+market}, namely unsupervised transfer learning based methods, and direct cross-dataset evaluation based methods. The first group of methods require images from the target dataset for unsupervised learning, which are not directly comparable to the second one that directly evaluates on the target dataset in consideration of real applications. The proposed QAConv method belongs to the second group. There are very few existing results of the same setting for the second group, except some baselines of other recent methods and the PN-GAN \cite{qian2018pose} which aims at augmenting source training data by GAN. For the comparison to the transfer learning methods, we consider that QAConv can serve as a better pre-trained model for them, and computing the RR+TLift on the fly is also more efficient than training on target dataset.
\begin{table}
    \centering
    \caption{Comparison of the state-of-the-art cross-dataset evaluation results (\%) with DukeMTMC-reID and Market-1501 as the target datasets.}\label{tab:duke+market}
    \begin{tabular}{|l||c|c|c|c||c|c|c|c|}
      \hline
      \multirow{2}{*}{\tabincell{c}{Method}}  & \multicolumn{2}{|c|}{Training}  & \multicolumn{2}{|c||}{Test: Duke}  & \multicolumn{2}{|c|}{Training}  & \multicolumn{2}{|c|}{Test: Market}\\
      \cline{2-9}
       & Source & Target & R1 & mAP & Source & Target & R1 & mAP \\
      \hline
      \hline
%      PTGAN, CVPR18~\cite{Wei-CVPR18-PTGAN} & Market & Duke & 27.4 & -& Duke & Market & 38.6 & -\\
%      SPGAN, CVPR18~\cite{deng2018image} & Market & Duke & 46.9 & 26.4& Duke & Market & 58.1 & 26.9\\
%      HHL, ECCV18~\cite{zhong2018generalizing} & Market & Duke & 46.9 & 27.2 & Duke & Market & 62.2 & 31.4\\
%      CamStyle, TIP19~\cite{zhong2018camstyle} & Market & Duke & 51.7 & 27.7 & Duke & Market & 64.7 & 30.4\\
%      %SyRI,	ECCV18~\cite{Bak-ECCV18-SyRI} & - & - & - & - & Multi & Market & 65.7 & -\\
%      CR-GAN+LMP, ICCV19~\cite{chen2019instance} & Market & Duke & 56.0 & 33.3 & Duke & Market & 64.5 & 33.2\\
%      %\hline
%      \hline
      PUL, TOMM18~\cite{fan2018unsupervised} & Market & Duke & 30.4 & 16.4& Duke & Market & 44.7 & 20.1\\					
      TJ-AIDL, CVPR18~\cite{wang2018transferable} & Market & Duke & 44.3 & 23.0 & Duke & Market & 58.2 & 26.5\\
      MMFA, BMVC18~\cite{lin2018multi} & Market & Duke & 45.3 & 24.7 & Duke &	Market & 56.7 & 27.4\\
      CFSM, AAAI19~\cite{chang2019disjoint} & Market & Duke & 49.8 & 27.3 & Duke & Market & 61.2 & 28.3\\
      DECAMEL, TPAMI19~\cite{Yu-TPAMI19-DECAMEL} &-&-&-&-& Multi & Market & 60.2 & 32.4\\
      PAUL, CVPR19~\cite{Yang-CVPR19-PAUL} & Market & Duke & 56.1 & 35.7 & Duke & Market & 66.7 & 36.8\\
      ECN, CVPR19~\cite{Zhong-CVPR19-ECN} & Market & Duke & 63.3 & 40.4 & Duke & Market & 75.1 & 43.0\\
      CDS, ICME19~\cite{Wu-ICME19-CDS} & Market & Duke & 67.2 & 42.7 & Duke & Market & 71.6 & 39.9\\
      %SSG, ICCV19 \cite{Fu_2019_ICCV} & Market & Duke & 73.0 & 53.4 & Duke & Market & 80.0 & 58.3 \\
      %\hline
      \hline
      ECN baseline, CVPR19~\cite{Zhong-CVPR19-ECN} & Market &  & 28.9 & 14.8 & Duke &  & 43.1 & 17.7\\
      PN-GAN, ECCV18~\cite{qian2018pose} & Market & & 29.9 & 15.8&-&&-&-\\
      QAConv$_{50}$ & Market & & 48.8 & 28.7  & Duke & & 58.6 & 27.2 \\
      QAConv$_{152}$ & Market & & 54.4 & 33.6  & Duke & & 62.8 & 31.6 \\
      QAConv$_{50}$ + RR + TLift & Market & & 64.5 & 55.1 & Duke & & 74.6 & 51.5  \\
      QAConv$_{152}$ + RR + TLift & Market & & 70.0 & 61.2 & Duke & & 78.7 & 58.2  \\
      \hline
      \hline
      MAR, CVPR19~\cite{Yu-CVPR19-MAR} & MSMT & Duke & 67.1 & 48.0 & MSMT & Market & 67.7 & 40.0\\
      PAUL, CVPR19~\cite{Yang-CVPR19-PAUL} & MSMT & Duke & 72.0 & 53.2 & MSMT & Market & 68.5 & 40.1\\
      %\hline
      \hline
      MAR baseline, CVPR19~\cite{Yu-CVPR19-MAR} & MSMT & & 43.1 & 28.8 & MSMT & & 46.2 & 24.6\\
      PAUL baseline, CVPR19~\cite{Yang-CVPR19-PAUL} & MSMT & & 65.7 & 45.6 & MSMT & & 59.3 & 31.0\\
      QAConv$_{50}$ & MSMT & & 69.4 & 52.6 & MSMT & & 72.6 & 43.1  \\
      QAConv$_{152}$ & MSMT & & 72.2 & 53.4 & MSMT & & 73.9 & 46.6  \\
      QAConv$_{50}$ + RR + TLift & MSMT & & 80.3 & 77.2  & MSMT & & 86.5 & 72.2\\
      QAConv$_{152}$ + RR + TLift & MSMT & & 82.2 & 78.4  & MSMT & & 88.4 & 76.0\\
      \hline
    \end{tabular}
\end{table}

\textbf{DukeMTMC-reID dataset.} As can be observed from Table \ref{tab:duke+market}, when trained on the Market-1501 dataset, QAConv achieves the best performance in the direct evaluation group with a large margin. When compared to transfer learning methods, QAConv also outperforms many of them except some very recent methods, indicating that QAConv enables the network to learn how to match two person images, and the learned model generalizes well in unseen domains without transfer learning. Besides, by enabling re-ranking and TLift, the proposed method achieves the best result among all except Rank-1 of CDS. Note that the re-ranking and TLift methods can also be incorporated into other methods, though. Therefore, we list their results separately. However, both of these are calculated on the fly without learning in advance, so together with QAConv, it appears that a ready-to-use method with good generalization ability can also be achieved even without further UDA, which is a nice solution considering that UDA requires heavy computation for deep learning in deployment phase.

When trained on MSMT17, QAConv itself beats all other methods except the transfer learning method PAUL. This is also the second best result among all existing methods taking DukeMTMC-reID as the target dataset, regardless of the training source. This clearly indicates QAConv's superiority in learning from large-scale data. It is preferred in practice in the sense that, when trained with large-scale data, there may be no need to adapt the learned model in deployment.

\textbf{Market-1501 dataset.} With Market-1501 as the target dataset as shown in Table \ref{tab:duke+market}, similarly, when trained with MSMT17, QAConv itself also achieves the best performance among others except Rank-1 of ECN. This can be considered a large advancement in cross-dataset evaluation, which is a better evaluation strategy for understanding the generalization ability of algorithms. Besides, when equipped with RR+TLift, the proposed method achieves the state of the art, with Rank-1 accuracy of 86.5\% and mAP of 72.2\%. Note that this comparison is not in a sense of \textit{fair}. We would like to share that beyond many recent efforts in UDA, enlarging the training data and exploiting on-the-fly computations in re-ranking and temporal fusion may also lead to good performance in unknown domain, with the advantage of no cost in training deep models everywhere.

\textbf{CUHK03 dataset.} The CUHK03 and MSMT17 datasets present large domain gaps to others. For CUHK03, it can be observed from Table \ref{tab:cuhk+msmt} that, with either Market-1501 or DukeMTMC-reID dataset as training set, QAConv without UDA performs better than a UDA method PUL~\cite{fan2018unsupervised}, and fairly comparable to another recent transfer learning method CDS \cite{Wu-ICME19-CDS}. However, all methods perform not well on the CUHK03 dataset. Only with the large MSMT17 data set as the source training data, the proposed method performs relatively better.

\textbf{MSMT17 dataset.} With the MSMT17 as target, only QAConv does not require adaptation in Table \ref{tab:cuhk+msmt}. However, it performs better than PTGAN \cite{Wei-CVPR18-PTGAN} and in part comparable to ECN \cite{Zhong-CVPR19-ECN}. This further confirms the generalizability of QAConv under large domain gaps, since without UDA it is already in part comparable to the state-of-the-art UDA methods. Note that TLift is not applicable on CUHK03 and MSMT17 due to no temporal information provided.
\begin{table}
    % \centering
    \caption{Comparison of the state-of-the-art cross-dataset evaluation results (\%) with CUHK03-NP (detected) and MSMT17 as the target datasets.}\label{tab:cuhk+msmt}
    \hspace{-4mm}
    \begin{tabular}{|l||c|c|c|c||c|c|c|c|}
      \hline
      \multirow{2}{*}{\tabincell{c}{Method}}  & \multicolumn{2}{|c|}{Training}  & \multicolumn{2}{|c||}{Test: CUHK03}  & \multicolumn{2}{|c|}{Training}  & \multicolumn{2}{|c|}{Test: MSMT}\\
      \cline{2-9}
       & Source & Target & R1 & mAP & Source & Target & R1 & mAP \\
      \hline
      \hline
%      DECAMEL, PAMI19~\cite{Yu-TPAMI19-DECAMEL} & - & -  & - & - & Multi & MSMT& 30.3 & 11.1\\
%      \hline
      PUL, TOMM18~\cite{fan2018unsupervised} & Market & CUHK03 & 7.6 & 7.3 & - & -  & - & - \\
      CDS, ICME19~\cite{Wu-ICME19-CDS} & Market & CUHK03 & 9.1 & 8.7 & - & -  & - & - \\
      PTGAN~\cite{Wei-CVPR18-PTGAN}, CVPR18 & - & -  & - & -  & Market & MSMT & 10.2 & 2.9\\
      ECN, CVPR19~\cite{Zhong-CVPR19-ECN} & - & -  & - & -  & Market & MSMT & 25.3 & 8.5\\
      \hline
      QAConv$_{50}$ & Market & & 9.9 & 8.6 & Market & & 22.6 & 7.0\\
      QAConv$_{152}$ & Market & & 14.1 & 11.8  & Market & & 25.6 & 8.2\\
      \hline
      \hline
      PUL, TOMM18~\cite{fan2018unsupervised} & Duke & CUHK03 & 5.6 & 5.2 & - & -  & - & - \\
      CDS, ICME19~\cite{Wu-ICME19-CDS} & Duke & CUHK03 & 8.1 & 7.1 & - & -  & - & - \\
      PTGAN~\cite{Wei-CVPR18-PTGAN}, CVPR18 & - & -  & - & -  & Duke & MSMT & 11.8 & 3.3\\
      ECN, CVPR19~\cite{Zhong-CVPR19-ECN} & - & -  & - & -  & Duke & MSMT & 30.2 & 10.2\\
      \hline
      QAConv$_{50}$ & Duke & & 7.9 & 6.8 & Duke & & 29.0 & 8.9\\
      QAConv$_{152}$ & Duke & & 11.0 & 9.4 & Duke & & 32.7 & 10.4\\
      \hline
      \hline
      QAConv$_{50}$ & MSMT & & 25.3 & 22.6 & -& -& -& -\\
      QAConv$_{152}$ & MSMT & & 32.6 & 28.1 &- & -& -& -\\
      \hline
    \end{tabular}
\end{table}

\subsection{Qualitative Analysis and Discussion}
\label{sec:discuss}
\begin{figure*}
\centering
\includegraphics[width=13mm]{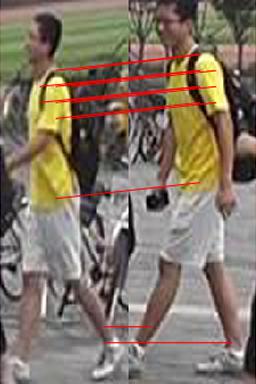} \hspace{1mm}
\includegraphics[width=13mm]{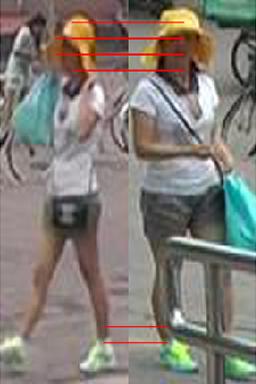}\hspace{1mm}
\includegraphics[width=13mm]{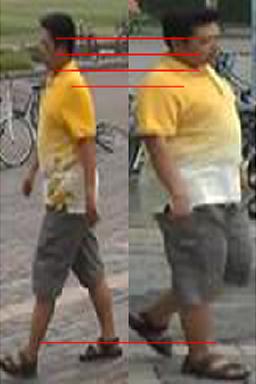}\hspace{1mm}
\includegraphics[width=13mm]{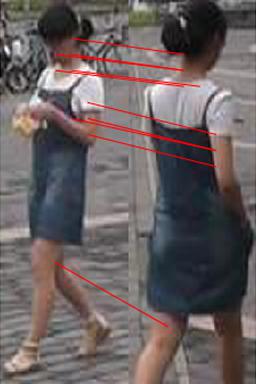}
\hspace{3mm}
\includegraphics[width=13mm]{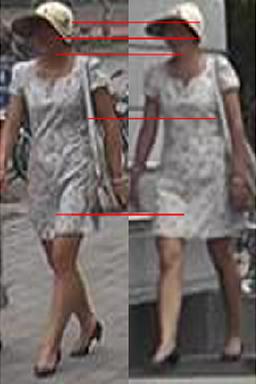}\hspace{1mm}
\includegraphics[width=13mm]{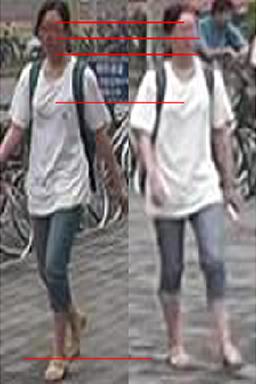}\hspace{1mm}
\includegraphics[width=13mm]{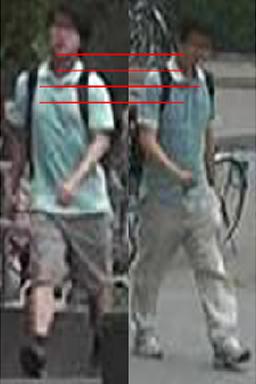}\hspace{1mm}
\includegraphics[width=13mm]{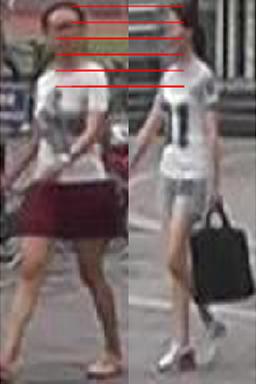}\\
0.71 \hspace{7mm} 0.63 \hspace{7mm} 0.62 \hspace{7mm} 0.66
\hspace{10mm}
0.73 \hspace{7mm} 0.67 \hspace{7mm} 0.57 \hspace{7mm} 0.58\\
(a) Positive pairs on Market-1501
\hspace{12mm}
(b) Negative pairs on Market-1501\\
\vspace{3mm}

\includegraphics[width=13mm]{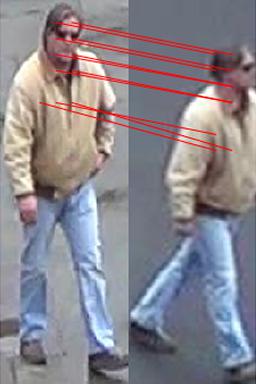}\hspace{1mm}
\includegraphics[width=13mm]{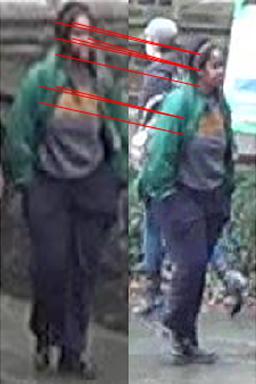}\hspace{1mm}
\includegraphics[width=13mm]{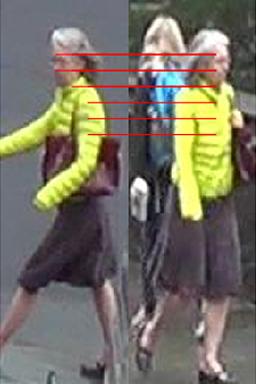}\hspace{1mm}
\includegraphics[width=13mm]{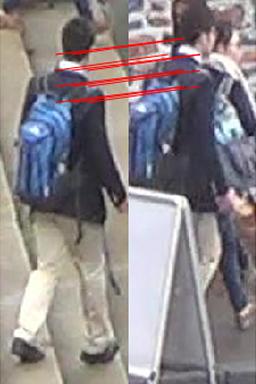}
\hspace{3mm}
\includegraphics[width=13mm]{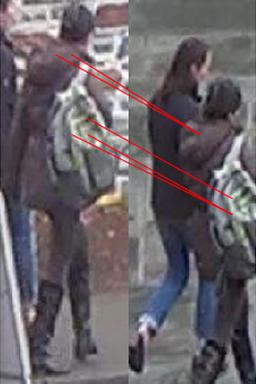}\hspace{1mm}
\includegraphics[width=13mm]{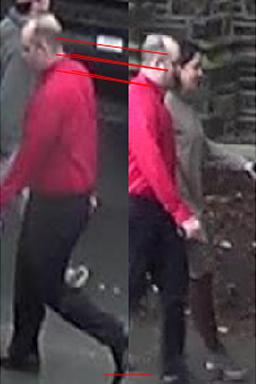}\hspace{1mm}
\includegraphics[width=13mm]{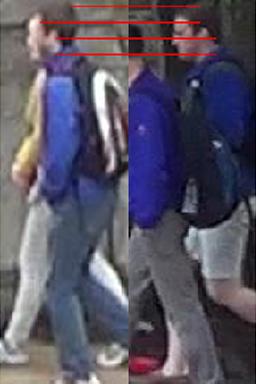}\hspace{1mm}
\includegraphics[width=13mm]{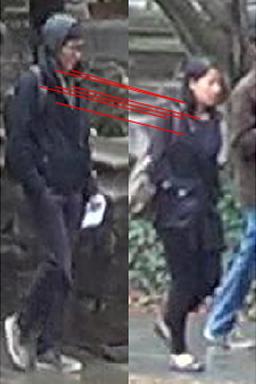}\\
0.64 \hspace{7mm} 0.62 \hspace{7mm} 0.66 \hspace{7mm} 0.60
\hspace{10mm}
0.54 \hspace{7mm} 0.58 \hspace{7mm} 0.54 \hspace{7mm} 0.50\\
(c) Positive pairs on DukeMTMC-reID
\hspace{5mm}
(d) Negative pairs on DukeMTMC-reID\\
%\vspace{2mm}
\caption{Examples of qualitative matching results by the proposed QAConv method using the model trained on the MSMT17 dataset. Numbers represent similarity scores.}\label{fig:samples}
\end{figure*}

A unique characteristic of the proposed QAConv method is its interpretation ability of the matching. Therefore, we show some qualitative matching results in Fig. \ref{fig:samples} for a better understanding of the proposed method. The model used here is trained on the MSMT17 dataset, and the evaluations are done on the query subsets of the Market-1501 and DukeMTMC-reID datasets. Results of both positive pairs and hard negative pairs are shown. Note that only reliable correspondences with matching scores over 0.5 are shown, and the local positions are coarse due to the $24\times8$ size of the feature map. As can be observed from Fig. \ref{fig:samples}, the proposed method is able to find correct local correspondences for positive pairs of images, even if there are notable misalignments in both scale and position, pose/viewpoint changes, occlusions, and mix up of other persons, thanks to the local matching mechanism of QAConv instead of global feature representations. Besides, for hard negative pairs, the matching of QAConv still appears to be mostly reasonable, by linking visually similar parts or even the same person (may be ambiguously labeled or walking closely to other persons). Note that the QAConv method gains the matching capability by automatic learning, from supervision of only class labels but not local correspondence labels.% Besides, the matching capability seems to be also generalizable to other datasets beyond the training set.

The QAConv network was trained on an NVIDIA DGX-1 server, with two V100 GPU cards. With the backbone network ResNet-50, the training time of QAConv on the DukeMTMC-reID dataset was 1.22 hours. In contrast, the most efficient softmax baseline took 0.72 hour for training. For deployment, the ECN~\cite{Zhong-CVPR19-ECN} reported 1 hour of transfer learning time with DukeMTMC-reID as target, while MAR~\cite{Yu-CVPR19-MAR} and DECAMEL~\cite{Yu-TPAMI19-DECAMEL} reported 10 and 35.2 hours of total learning time, respectively, compared to the ready-to-use QAConv. For inference, with the DukeMTMC-reID dataset as target, QAConv took 26 seconds for feature extraction and 26 seconds for similarity computation. In contrast, the softmax baseline took 26 seconds for feature extraction and 0.2 seconds for similarity computation. Besides, the proposed method took 303 seconds for reranking, and 67 seconds for TLift. This is still efficient, especially for RR+TLift, compared to transfer learning for deployment. Therefore, the overall solution of QAConv+RR+TLift is promising in practical applications.

%One drawback of QAConv is that it requires more memory to run than other methods, and it needs to store feature maps of images, rather than features, where feature maps are generally larger in size than representation features. For training on the DukeMTMC-reID, the GPU memory consumption for the QAConv is about 2.83GB, while that for the softmax baseline is about 2.78GB. They are comparable because though QAConv spends some more on class memory, it uses three layers of the ResNet-50, while the softmax baseline uses four layers. For inference, the peak GPU memory for the QAConv is about 2.3GB, while that for the softmax baseline is about 1.7GB.

For further analysis on memory usage, please see the Appendix. As for the TLift, it can only be applied on datasets with good time records. Though this information is easy to obtain in real surveillance, most existing person re-identification datasets do not contain it. Another drawback of TLift is that it cannot be applied to arbitrary query images beyond a camera network, though once an initial match is found, it can be used to refine the search. Besides, it cannot help when there is no nearby person with the query.

\section{Conclusion}
In this paper, through extensive experiments we show that the proposed QAConv method is quite promising for person matching without further transfer learning, and it has a much better generalization ability than existing baselines. Though QAConv can also be plugged into other transfer learning methods as a better pre-trained model, in practice, according to the experimental results of this paper, we suggest a ready-to-use solution which works in the following principles. First, a large-scale and diverse training data (e.g. MSMT17) is required to learn a generalizable model. Second, a larger network (e.g. ResNet-152) benefits a better overall performance, which could be further distilled into smaller ones for efficiency. Finally, score re-ranking and temporal fusion model such as TLift can be computed on the fly in deployment, which can largely improve performance and they are more efficient to use than transfer learning.

\section*{Acknowledgements}
This work was partly supported by the NSFC Project \#61672521. The authors would like to thank Yanan Wang who helped producing several illustration figures in this paper, Jinchuan Xiao who optimized the TLift code, and Anna Hennig who helped proofreading the paper.

\clearpage
\bibliographystyle{splncs04}
\bibliography{../../Bib/Liao}

% \linenumbers
\pagestyle{headings}
\mainmatter
\def\ECCVSubNumber{1369}  % Insert your submission number here

\title{Appendix} % Replace with your title

\begin{comment}
\titlerunning{ECCV-20 submission ID \ECCVSubNumber}
\authorrunning{ECCV-20 submission ID \ECCVSubNumber}
\author{Anonymous ECCV submission}
\institute{Paper ID \ECCVSubNumber}
\end{comment}
%******************

%\begin{comment}
\titlerunning{ECCV 2020: Person Re-Identification with QAConv and TLift}
\author{}
\authorrunning{Shengcai Liao and Ling Shao}
\institute{}
%\end{comment}
%******************
\maketitle

%%%%%%%%% BODY TEXT

\section{Role of Random Occlusion}

The random occlusion (RO) we designed for data augmentation is similar to the random erasing (RE) \cite{zhong2020random} and cutout \cite{devries2017improved} methods. In the RE implementation, the target erasing area is sampled from a combination of random area and aspect ratio, which could exceed the original image height or width. Therefore, it needs to try multiple times (100 by default) to generate a reasonable region for erasing. In contrast, in our implementation of the random occlusion, a square area is used, with the size randomly sampled at most $0.8\times width$ of the image, and randomly put in a valid location. Then the square area is filled with white pixels. Note that with a simple square area, there is no need to sample multiple times of areas and aspect ratios and check the validity, and hence the generation process is more efficient. As for the cutout method, it uses multiple square regions in fixed sizes specified by hyperparameters, but not in random. The fixed-size regions may make the cut either too small or too large, and so it is not very convenient to set.

To show their differences, in the training of QAConv, we compare these data augmentation methods as well as a baseline without any random occlusion. From the results shown in Table \ref{tab:occlusion}, it can be observed that the three data augmentation methods generally improve the baseline which does not apply any random occlusion. Intuitively, they are useful for QAConv because random occlusion forces QAConv to learn various local correspondences, instead of only salient but easy ones. Besides, the three data augmentation methods perform comparable, with the RO implementation being slightly better. Therefore, considering also the efficiency of the RO implementation, it is adopted in the training of the proposed QAConv algorithm.

\begin{table}
    \centering
    \caption{Role of random occlusion.}\label{tab:occlusion}
    \begin{tabular}{|l|c|c|c|c|}
      \hline
      \multirow{2}{*}{\tabincell{c}{Method}}  & \multicolumn{2}{|c|}{Market$\rightarrow$Duke} &  \multicolumn{2}{|c|}{Duke$\rightarrow$Market} \\
      \cline{2-5}
        & Rank-1 & mAP  & Rank-1 & mAP \\
      \hline
      QAConv without occlusion & 50.5 & 29.5 & 61.6 & 28.4\\
      QAConv with RE~\cite{zhong2020random} & 51.6 & 30.6 & 62.0 & 29.8\\
      QAConv with cutout~\cite{devries2017improved} & 51.6 & 30.8 & 62.6 & 30.3\\
      QAConv with RO & \textbf{54.4} & \textbf{33.6} &\textbf{62.8} &\textbf{31.6}\\
      \hline
    \end{tabular}
\end{table}

\section{Complete Comparisons of Backbone Networks}
Tables \ref{tab:duke} and \ref{tab:market} show complete comparisons between the QAConv results with the ResNet-50 as backbone (denoted as QAConv$_{50}$) and with the ResNet-152 as backbone (denoted as QAConv$_{152}$), with DukeMTMC-reID and Market-1501 as the target datasets, respectively. Results of applying re-ranking alone are not shown in the main paper.
\begin{table}
    \centering
    \caption{Comparison (\%) of backbone networks with DukeMTMC-reID as the target dataset.}\label{tab:duke}
    \begin{tabular}{|l|c|c|c|c|}
      \hline
      \multirow{2}{*}{\tabincell{c}{Method}}  & \multicolumn{2}{|c|}{Training}  & \multicolumn{2}{|c|}{Test: Duke}\\
      \cline{2-5}
       & Source & Target & R1 & mAP \\
      \hline
      \hline
      QAConv$_{50}$ & Market & & 48.8 & 28.7\\
      QAConv$_{152}$ & Market & & 54.4 & 33.6 \\
      QAConv$_{50}$ + RR & Market & & 56.9 & 47.8 \\
      QAConv$_{152}$ + RR & Market & & 61.8 & 52.4 \\
      QAConv$_{50}$ + RR + TLift & Market & & 64.5 & 55.1 \\
      QAConv$_{152}$ + RR + TLift & Market & & 70.0 & 61.2 \\
      \hline
      \hline
      QAConv$_{50}$ & MSMT & & 69.4 & 52.6\\
      QAConv$_{152}$ & MSMT & & 72.2 & 53.4\\
      QAConv$_{50}$ + RR & MSMT & & 76.7 & 71.2 \\
      QAConv$_{152}$ + RR & MSMT & & 78.1 & 72.4 \\
      QAConv$_{50}$ + RR + TLift & MSMT & & 80.3 & 77.2 \\
      QAConv$_{152}$ + RR + TLift & MSMT & & 82.2 & 78.4\\
      \hline
    \end{tabular}
\end{table}
\begin{table}
    \centering
    \caption{Comparison (\%) of backbone networks with Market-1501 as the target dataset.}\label{tab:market}
    \begin{tabular}{|l|c|c|c|c|}
      \hline
      \multirow{2}{*}{\tabincell{c}{Method}}  & \multicolumn{2}{|c|}{Training}  & \multicolumn{2}{|c|}{Test: Market}\\
      \cline{2-5}
       & Source & Target & R1 & mAP \\
      \hline
      \hline
      QAConv$_{50}$ & Duke & & 58.6 & 27.2 \\
      QAConv$_{152}$  & Duke & & 62.8 & 31.6 \\
      QAConv$_{50}$ + RR & Duke & & 65.7 & 45.8 \\
      QAConv$_{152}$ + RR & Duke & & 68.5 & 51.2 \\
      QAConv$_{50}$ + RR + TLift & Duke & & 74.6 & 51.5  \\
      QAConv$_{152}$ + RR + TLift & Duke & & 78.7 & 58.2  \\
      \hline
      \hline
      QAConv$_{50}$  & MSMT & & 72.6 & 43.1  \\
      QAConv$_{152}$  & MSMT & & 73.9 & 46.6  \\
      QAConv$_{50}$ + RR   & MSMT & & 77.4 & 65.6 \\
      QAConv$_{152}$ + RR  & MSMT & & 79.2 & 69.1 \\
      QAConv$_{50}$ + RR + TLift   & MSMT & & 86.5 & 72.2\\
      QAConv$_{152}$ + RR + TLift  & MSMT & & 88.4 & 76.0\\
      \hline
    \end{tabular}
\end{table}

\section{Comparisons to Other Losses}

Since the loss of hard triplet mining \cite{hermans2017defense} is popular in person re-identification, we further include it in the loss comparisons. Besides, we provide a further analysis on different loss configurations of the QAConv. The results are shown in Table \ref{tab:loss} under Market$\rightarrow$Duke, where triplet results are each with its best margin. While the mini-batch hard triplet loss does improve the softmax cross-entropy loss, it seems that it is not efficient in learning the QAConv, possibly because local matching requires large pairs to learn, as done with the proposed class memory and focal loss, but not in mini-batches. Note that focal loss is a bit aggressive in learning, but softly. However, the hard triplet loss is in fact more aggressive.

\begin{table}
    \centering
    \caption{Role of loss functions under Market$\rightarrow$Duke (\%).}\label{tab:loss}
    \begin{tabular}{|c|c|c|c|}
      \hline
      \multicolumn{2}{|c|}{Method} & Rank-1 & mAP \\
      \hline
      \multirow{5}{*}{\tabincell{c}{ResNet-152}} & Softmax cross-entropy & 34.9 & 18.4\\
      & Softmax cross-entropy + triplet & 39.6 & 23.0\\
      & Arc loss~\cite{Deng-CVPR19-ArcFace} & 35.3 & 17.1\\
      & Center loss~\cite{Wen-ECCV16-CenterLoss,Jin-IJCB17-CenterLossReID} & 38.9 & 22.1\\
      & Class memory loss & 40.7 & 21.8\\
      \hline
      \multirow{7}{*}{\tabincell{c}{QAConv$_{50}$}} & Mini-batch triplet (w/o class memory) & 42.2 & 23.7\\
      & Softmax cross-entropy & 43.4 & 24.9\\
      & Binary cross-entropy & 46.1 & 27.3\\
      & Softmax cross-entropy + triplet & 44.3 & 24.2 \\
      & Binary cross-entropy + triplet & 44.7 & 23.6\\
      & Focal loss + triplet & 43.3 & 23.2\\
      & Focal loss (default) & \textbf{48.8} & \textbf{28.7}\\
      \hline
    \end{tabular}
\end{table}

\section{Fusion of Global Similarity}

To see whether fusing a global similarity branch helps improving the performance, we tried an extra global feature learning branch by performing a global average pooling on the final feature maps, and a softmax cross-entropy loss for classification. During testing, the cosine similarity computed from this global feature branch is fused to the QAConv similarity. However, after trying different weights of the two losses, the best mAP we can get is 28.4\% under Market$\rightarrow$Duke, with the weight 0.001 of the global branch. It is a bit worse than the default QAConv (28.7\%). This may be because the vanilla global feature branch cannot handle misalignments and occlusions, and so more advanced techniques are needed here. This deserves a further study.

\section{TLift for Other Methods}

Note that TLift can also be generally applied to other methods for improvements. To demonstrate this, Tables \ref{tab:tlift-market} and \ref{tab:tlift-duke} show results of applying TLift to all baseline methods under Market$\rightarrow$Duke and Duke$\rightarrow$Market, respectively. It can be observed that, beyond the improvements made by re-ranking, TLift can further improve all baseline methods. The improvements are consistently large, with Rank-1 improved by 10.1\%-14.1\%, and mAP improved by 3.6\%-11.1\%.

\begin{table}
    \centering
    \caption{Role of TLift under Market$\rightarrow$Duke (\%).}\label{tab:tlift-market}
    \begin{tabular}{|c|c|c|c|c|c|c|}
      \hline
      \multirow{2}{*}{\tabincell{c}{Method}}  & \multicolumn{2}{|c|}{Original} &  \multicolumn{2}{|c|}{+ RR} & \multicolumn{2}{|c|}{+ RR + TLift}\\
      \cline{2-7}
        & Rank-1 & mAP  & Rank-1 & mAP & Rank-1 & mAP\\
      \hline
      Softmax cross-entropy & 34.9 & 18.4 & 41.5 & 30.5 & 51.7 & 39.7\\
      Arc loss~\cite{Deng-CVPR19-ArcFace} & 35.3 & 17.1 & 39.8 & 26.3 & 51.0 & 34.8\\
      Center loss~\cite{Wen-ECCV16-CenterLoss,Jin-IJCB17-CenterLossReID} & 38.9 & 22.1 & 42.5 & 31.5 & 56.6 & 42.6\\
      Class memory loss & 40.7 & 21.8 & 47.8 & 36.1 & 59.6 & 46.2\\
      QAConv & \textbf{54.4} & \textbf{33.6} & \textbf{61.8} & \textbf{52.4} & \textbf{70.0} & \textbf{61.2}\\
      \hline
    \end{tabular}
\end{table}

\begin{table}
    \centering
    \caption{Role of TLift under Duke$\rightarrow$Market (\%).}\label{tab:tlift-duke}
    \begin{tabular}{|c|c|c|c|c|c|c|}
      \hline
      \multirow{2}{*}{\tabincell{c}{Method}}  & \multicolumn{2}{|c|}{Original} &  \multicolumn{2}{|c|}{+ RR} & \multicolumn{2}{|c|}{+ RR + TLift}\\
      \cline{2-7}
        & Rank-1 & mAP  & Rank-1 & mAP & Rank-1 & mAP\\
      \hline
      Softmax cross-entropy & 48.5 & 21.4 & 53.2 & 33.7 & 63.3 & 38.0\\
      Arc loss~\cite{Deng-CVPR19-ArcFace} & 48.9 & 21.4 & 54.5 & 34.8 & 64.8 & 39.3\\
      Center loss~\cite{Wen-ECCV16-CenterLoss,Jin-IJCB17-CenterLossReID} & 48.8 & 22.0 & 52.5 & 33.3 & 63.0 & 36.9\\
      Class memory loss & 47.8 & 20.5 & 52.9 & 33.1 & 63.4 & 37.5\\
      QAConv & \textbf{62.8} & \textbf{31.6} & \textbf{68.5} & \textbf{51.2} & \textbf{78.7} & \textbf{58.2}\\
      \hline
    \end{tabular}
\end{table}

\begin{table}
    \centering
    \caption{Influence of TLift parameters under Market$\rightarrow$Duke (\%). Bold numbers are with the default parameters.}\label{tab:tlift-para-market}
    \linespread{1.2}\selectfont
    \setlength{\tabcolsep}{2mm}{
    \begin{tabular}{|c|c c c c c c c c c c|}
      \hline
      $\tau$ & 50 & \textbf{100}	& 150 &	200	& 250 &	300	& 350 &	400	& 450 &	500\\
      \hline
      Rank-1 & 69.3 & \textbf{70.0} & 69.7 & 69.8 & 69.1 & 68.3 & 66.8 & 65.5 & 64.4 & 63.9 \\
      mAP & 60.7 & \textbf{61.2} & 60.7 & 59.9 & 58.8 & 57.3 & 55.7 & 54.0 & 52.4 & 51.2\\
      \hline
    \end{tabular}
    \\[5mm]
    \begin{tabular}{|c|c c c c c c c c c c|}
      \hline
      $\sigma$ & 50 & 100 & 150 & \textbf{200}	& 250 &	300	& 350 &	400	& 450 &	500\\
      \hline
      Rank-1 & 67.4 & 69.5 & 70.4 & \textbf{70.0} & 69.4 & 69.2 & 68.9 & 68.4 & 68.0 & 67.7\\
      mAP & 55.4 & 59.6 & 60.9 & \textbf{61.2} & 61.0 & 60.8 & 60.5 & 60.1 & 59.8 & 59.5 \\
      \hline
    \end{tabular}
    \\[5mm]
    \begin{tabular}{|c|c c c c c c c c c c|}
      \hline
      $K$ & 5 & \textbf{10} & 15 & 20 & 30 & 40 & 50 & 100 & 150 & 200\\
      \hline
      Rank-1 & 69.7 & \textbf{70.0} & 70.2 & 70.0 & 69.4 & 68.9 & 68.2 & 67.0 & 65.5 & 64.8\\
      mAP & 60.8 & \textbf{61.2} & 61.2 & 61.0 & 60.3 & 59.6 & 58.8 & 56.8 & 55.7 & 55.2\\
      \hline
    \end{tabular}
    \\[5mm]
    \begin{tabular}{|c|c c c c c c c c c c|}
      \hline
      $\alpha$ & 0.01 & 0.02 & 0.05 & 0.1 & \textbf{0.2} & 0.3 & 0.4 & 0.5 & 0.7 & 1\\
      \hline
      Rank-1 & 70.4 & 70.4 & 70.2 & 70.2 & \textbf{70.0} & 69.4 & 69.1 & 68.6 & 68.3 & 67.5 \\
      mAP & 60.8 & 60.8 & 60.8 & 61.0 & \textbf{61.2} & 61.1 & 61.0 & 60.9 & 60.4 & 59.7\\
      \hline
    \end{tabular}}
\end{table}

\section{Parameter Analysis}

Considering the memory consumption and the efficiency, the kernel size of QAConv is set to $s=1$. Parameters for TLift are $\tau=100$, $\sigma=200$, $K=10$, and $\alpha=0.2$. They were fixed in all experiments after some initial tries. To understand their influence, we vary them one by one, with corresponding results shown in Tables \ref{tab:tlift-para-market} and \ref{tab:tlift-para-duke}. It can be observed that, the parameters are not sensitive in a broad range, so that they are easy to select. Besides, some better results can be obtained by varying parameters other than the defaults.

\begin{table}
    \centering
    \caption{Influence of TLift parameters under Duke$\rightarrow$Market (\%). Bold numbers are with the default parameters.}\label{tab:tlift-para-duke}
    \linespread{1.2}\selectfont
    \setlength{\tabcolsep}{2mm}{
    \begin{tabular}{|c|c c c c c c c c c c|}
      \hline
      $\tau$ & 50 & \textbf{100}	& 150 &	200	& 250 &	300	& 350 &	400	& 450 &	500\\
      \hline
      Rank-1 & 76.2 & \textbf{78.7} & 79.8 & 79.7 & 79.9 & 79.0 & 78.6 & 78.2 & 77.6 & 77.2 \\
      mAP & 57.2 & \textbf{58.2} & 58.6 & 58.4 & 58.2 & 57.7 & 57.2 & 56.6 & 56.0 & 55.4\\
      \hline
    \end{tabular}
    \\[5mm]
    \begin{tabular}{|c|c c c c c c c c c c|}
      \hline
      $\sigma$ & 50 & 100	& 150 &	\textbf{200}	& 250 &	300	& 350 &	400	& 450 &	500\\
      \hline
      Rank-1 & 76.1 & 78.5 & 78.6 & \textbf{78.7} & 78.6 & 78.1 & 78.0 & 77.9 & 77.9 & 77.6 \\
      mAP & 55.6 & 57.6 & 58.1 & \textbf{58.2} & 58.5 & 58.7 & 58.8 & 59.0 & 59.1 & 59.2\\
      \hline
    \end{tabular}
    \\[5mm]
    \begin{tabular}{|c|c c c c c c c c c c|}
      \hline
      $K$ & 5 & \textbf{10} & 15 & 20 & 30 & 40 & 50 & 100 & 150 & 200\\
      \hline
      Rank-1 & 79.6 & \textbf{78.7} & 78.1 & 77.6 & 76.6 & 76.2 & 75.8 & 74.4 & 73.4 & 72.7\\
      mAP & 56.9 & \textbf{58.2} & 58.4 & 58.3 & 58.0 & 57.9 & 57.8 & 57.3 & 56.6 & 55.9\\
      \hline
    \end{tabular}
    \\[5mm]
    \begin{tabular}{|c|c c c c c c c c c c|}
      \hline
      $\alpha$ & 0.01 & 0.02 & 0.05 & 0.1 & \textbf{0.2} & 0.3 & 0.4 & 0.5 & 0.7 & 1\\
      \hline
      Rank-1 & 78.4 & 78.5 & 78.7 & 78.8 & \textbf{78.7} & 78.5 & 78.0 & 77.6 & 76.5 & 75.4\\
      mAP &  53.8 & 54.1 & 55.0 & 56.3 & \textbf{58.2} & 59.4 & 59.9 & 60.0 & 59.5 & 58.5\\
      \hline
    \end{tabular}}
\end{table}

\section{Memory Usage}

One drawback of QAConv is that it requires more memory to run than other methods, and it needs to store feature maps of images, rather than features, where feature maps are generally larger in size than representation features. For training on the DukeMTMC-reID, the GPU memory consumption for the QAConv is about 2.83GB, while that for the softmax baseline is about 2.78GB. They are comparable because though QAConv spends some more on class memory, it uses three layers of the ResNet-50, while the softmax baseline uses four layers. For inference, the peak GPU memory for the QAConv is about 2.3GB, while that for the softmax baseline is about 1.7GB.

\section*{Biography}

Shengcai Liao is a Lead Scientist in the Inception Institute of Artificial Intelligence (IIAI), Abu Dhabi, UAE. He is a Senior Member of IEEE. Previously, he was an Associate Professor in the Institute of Automation, Chinese Academy of Sciences (CASIA). He received the B.S. degree in mathematics from the Sun Yat-sen University in 2005 and the Ph.D. degree from CASIA in 2010. He was a Postdoc in the Michigan State University during 2010-2012. His research interests include object detection, face recognition, and person re-identification. He has published over 100 papers, with over 11,000 citations according to Google Scholar. He was awarded the Best Student Paper in ICB 2006, ICB 2015, and CCBR 2016, and the Best Paper in ICB 2007. He was also awarded the IJCB 2014 Best Reviewer and CVPR 2019 Outstanding Reviewer. He was an Assistant Editor for the book “Encyclopedia of Biometrics (2nd Ed.)”. He also served as Area Chairs for ICPR 2016, ICB 2016, and ICB 2018, and reviewers for ICCV, CVPR, ECCV, TPAMI, IJCV, TIP, TIFS, etc. He was the Winner of the CVPR 2017 Detection in Crowded Scenes Challenge and the ICCV 2019 NightOwls Pedestrian Detection Challenge. Homepage: \url{https://liaosc.wordpress.com/}

~\\

Ling Shao (Senior Member, IEEE) is currently the Executive Vice President and a Provost of the Mohamed bin Zayed University of Artificial Intelligence. He is also the CEO and the Chief Scientist of the Inception Institute of Artificial Intelligence (IIAI), Abu Dhabi, United Arab Emirates. His research interests include computer vision, machine learning, and medical imaging. He is a fellow of IAPR, IET, and BCS.

% \clearpage
\bibliographystyle{splncs04}
\bibliography{../../Bib/Liao}

\end{document}